# Object Classification in Images of Neoclassical Artifacts Using Deep Learning

**Bernhard Bermeitinger**
bernhard.bermeitinger@uni-passau.de
Universität Passau, Germany

**Simon Donig**
simon.donig@uni-passau.de
Universität Passau, Germany

**Maria Christoforaki**
maria.christoforaki@uni-passau.de
Universität Passau, Germany

**André Freitas**
andre.freitas@uni-passau.de
Universität Passau, Germany

**Siegfried Handschuh**
siegfried.handschuh@uni-passau.de
Universität Passau, Germany

## Classifying aesthetic forms – a methodology at the heart of art history

The transformation of aesthetic styles has been at the heart of art history since its inception as a scholarly discipline in the late eighteenth century. Analyzing the single artifact and the carefully curated corpus have been the techniques for crafting hermeneutic understanding for such processes of change. Recently new instruments based on statistical techniques empower us for a fresh take on bodies of sources once disregarded as second tier complementary sources such as for instance very large corpora.

## The Neoclassica research framework

The *Neoclassica* research framework (Donig et al., 2016) was conceived to provide scholars with new instruments and methods for analyzing and classifying artifacts and aesthetic forms from the era of Classicism (ca. 1760–1860). The neoclassic movement was of almost global scale—affecting architecture and design from Sidney to New York, and from Athens to the outreach of the Russian Urals—while relating to a common reference in classical antiquity, therefore making it an ideal topic for studying processes of stylistic transformation.

It accommodates both traditional knowledge representation as a formal ontology and data-driven knowledge discovery, where cultural patterns will be identified by means of algorithms in statistical analysis and machine learning, having in particular the potential to uncover hitherto unknown patterns in the source data. The outcomes of the top-down and the bottom-up approach will be united in a consistent, unified formal knowledge representation.

Motivated by the need to combine object classification with domain knowledge representation, the ontology focuses at the moment on artifacts (in particular furniture and architecture) and their components. Following the preliminary hypotheses that aesthetic forms in furniture and architecture are in closest communication with each other due to constructional commonalities and their shared reference of the Classic, we decided to start developing the knowledge discovery module of *Neoclassica* by classifying artifacts in digital images.

## Knowledge discovery

In this paper, we report on our efforts for using deep learning for classifying artifacts in digital visuals. We chose a deep learning approach for our classification method because of its current superiority over other methods and still rising accuracy over the last years in nearly all image classification and object detection challenges.

Initially, we compiled a body of images both from commercial sources such as auction houses, antique dealers and other public sources. Due to the complex copyright situation, this corpus can not be redistributed. In order to make our experiment reproducible and since the *Metropolitan Museum of Art* (MET) has released 375,000 images in the public domain (The Metropolitan Museum of Art, 2017). we assembled a corpus of 379 artifacts relevant to our research. We processed this corpus with the same algorithm as the original proprietary corpus and released the data together with the source code (The Neoclassica Project, 2017).

### Classifier description

The main classifier for our experiments is a *Convolutional Neural Network* (CNN). It classifies an input image as a whole.

In a first step, we applied a standard implementation of a CNN (namely *VGG19* (Russakovsky et al., 2015)). The results were not satisfactory for our needs. It classified the type of the object depicted in the image with an accuracy of 0.37.

In a second step, we opted to employ pre-training, a common technique for improving accuracy in neural networks. We experienced that available pre-trained classifiers for generic image classification proved ineffective in our case. Most of them are trained on a specific subset of *ImageNet* (Deng et al., 2009), containing 1000 classes. These classes are broadly spread around everyday objects like dogs and planes. This led us to assume that the amount of very different classes that don't occur in our corpora interfere with the classification. Following that hypothesis, we decided to train the algorithm on a specific subset compiled from *ImageNet* mainly containing different furniture objects like tables, chairs, and cabinets. They sum up to 35,000 images. The first training step with these images resulted in an accuracy of 0.54 of classifying the object correctly.

### First layout

The first corpus contained 2,129 images representing 300 European period artifacts mostly in a colored format of highly diverging quality and resolution. They depict the objects fully, partially or are close-up shots of specific forms. We coarsely annotated these images by manually labeling them on the level of folders. The concepts applied during this labeling process are directly taken from the *Neoclassica* ontology and describe concepts for types of artifacts. These concepts were derived from period sources.

The depth of the class hierarchy was partly reflected by the folder structure. The folder labeled "Chest of drawers" contains all instances of this class. Their labels in turn reflect the names of all the subclasses in the most extensive specification (e.g. semainier, Wellington chest, commode scriban).

After pre-training, the next step was fine-tuning with this corpus. The accuracy was 0.44, the F1 measure 0.44.

### Second layout

The second corpus was assembled from open data released by the MET. It contains 1,246 images representing 379 European and American period artifacts ranging roughly from 1780–1840 including some transition pieces, drawings, and prints. They also depict the objects fully, partially, or are close-up shots of specific forms. We used the titles provided by the MET and manually aligned them with the *Neoclassica* ontology.

The overall mean accuracy over all classes was 0.36, the F1 measure 0.21. For the computation of these numbers, all results that are non-computable (due to only having one image in either the train or test set) were removed. These low numbers result from the existence of two many artifacts represented by only one image, thus making a split in training and testing data meaningless. However, applying pre-training using same *ImageNet* corpus as in the first layout yielded a mean accuracy over all classes of 0.59 and a F1 measure of 0.58.

In order to achieve better results and since the classifier classifies the image as whole, we excluded all images that did not depict the whole artifact. We kept multiple copies of the same image if they were used to describe a different but similar object. We split the images depicting multiple objects so that the resulting images represent only one artifact. We also processed these images so that neighboring objects were covered with solid colors. The images that could not be split (e.g. room interiors) were excluded from the corpus.

Using the same settings with the curated corpus and with pre-training we achieved an accuracy of 0.77 and an F1 measure of 0.76.

### Challenges

While pre-training and improving the curation process helped us to raise the accuracy, we assume that there is room for improvement.

Parameters to be taken into consideration include the small size of corpora and how to overcome this limitation since this limits the effectiveness of a neural net. Additionally, since pre-training has been proven to enhance the results, it is rational to assume that a pre-training corpus better suited to period artifacts would improve the results further. Third, our experiment was affected by the limitation of the standard implementation of the CNN which classifies the image as a whole and not parts of it.

Outlining parts inside an image and classifying them is a difficult task for machine learning methods. Recently, a new type of neural net emerged that tackles this challenge: *Regional CNN* (RCNN). It is implemented most prominently in an algorithm called *MultiPath Network* (Zagoruyko et al., 2016).

4Current improvements: Using a Regional CNN

We are manually annotating regions within the images with classes from the ontology for training a RCNN that locates objects.

The implementation of the RCNN is divided into two steps. First, it detects objects in the image and draws their outline as a polygon. The second step is classifying the outlined objects using the included standard CNN.

Preliminary results of *MultiPath Network* with default pre-trained settings show that the first step of the RCNN already outlines objects in our corpora within reasonable limits. The corpus for pre-training is *COCO* (Lin et al., 2014). Naturally, specific domain objects are not located and the class names are too generic. For our purpose, fine-tuning on a custom annotated corpus is essential. An RCNN requires a more detailed corpus. The exhaustive task of manually draw the objects' outlines within an image promises higher quality in locating objects (first step) and is necessary to classify the objects according to the ontology (second step).

### Future work

We decided to take two steps in the near future for improving our results.

First, we are compiling a new corpus to train the RCNN with, avoiding pitfalls like inconsistent quality, heterogeneous image rights and an inadequate distribution of image per class. Here we would like to go a dual approach. Together with domain experts, we intend to collate a corpus from the large repository of a major auction house, providing us not only with a selection of artifacts' images but also with texts to be used in multimodal analysis.

On the other hand, this kind of artifacts may exhibit provenance issues (e.g. heterogeneity or lack of provenance). We will thus compensate for such issues by digitizing a major corpus of neoclassical artifacts forming an ensemble and comprising artifacts in multiple modes having evolved in close reference to each other. Therefore, we have entered a partnership with the Dessau-Wörlitz UNESCO world-heritage site, an almost untouched complex of manor houses and their furnishings in early neoclassical style.

Regarding the annotations, we are developing our own semantic annotation and ontology population tool since January 2017. The tool will create an annotated corpus. The actual annotation process will be conducted in cooperation with emerging domain experts from the chair of Visual Culture and Art History at the University Passau.